%% file: main.tex
\documentclass[sigconf]{acmart}
\usepackage[ruled]{algorithm2e}
\usepackage{comment}
\usepackage{subcaption}
\usepackage{hyperref}
\usepackage{caption} 
\usepackage{enumitem}
\usepackage{multirow}

\AtBeginDocument{%
  \providecommand\BibTeX{{%
    \normalfont B\kern-0.5em{\scshape i\kern-0.25em b}\kern-0.8em\TeX}}}

\copyrightyear{2023}
\acmYear{2023}
\setcopyright{rightsretained}
\acmConference[WSDM '23]{Proceedings of the Sixteenth ACM International Conference on Web Search and Data Mining}{February 27-March 3, 2023}{Singapore, Singapore}
\acmBooktitle{Proceedings of the Sixteenth ACM International Conference on Web Search and Data Mining (WSDM '23), February 27-March 3, 2023, Singapore, Singapore}\acmDOI{10.1145/3539597.3570443}
\acmISBN{978-1-4503-9407-9/23/02}

\begin{document}

\newcommand{\Require}{\textbf{Input:}\,}
\newcommand{\Ensure}{\textbf{Output:}\,}
\newcommand{\model}{{MetaCRS}}

\newcommand\hnote[1]{\textcolor{red}{[Wang]: #1}}
\newcommand\cnote[1]{\textcolor{blue}{[Chu]: #1}}

\title{Meta Policy Learning for Cold-Start Conversational Recommendation}


\begin{abstract}
Conversational recommender systems (CRS) explicitly solicit users' preferences for improved recommendations on the fly. Most existing CRS solutions count on a single policy trained by reinforcement learning for a population of users. However, for users new to the system, such a global policy becomes ineffective to satisfy them, i.e., the cold-start challenge. In this paper, we study CRS policy learning for cold-start users via meta reinforcement learning. We propose to learn a meta policy and adapt it to new users with only a few trials of conversational recommendations. To facilitate fast policy adaptation, we design three synergetic components. Firstly, 
we design a meta-exploration policy dedicated to identifying user preferences via a few exploratory conversations, which accelerates personalized policy adaptation from the meta policy.  Secondly, we adapt the item recommendation module for each user to maximize the recommendation quality based on the collected conversation states during conversations. 
Thirdly, we propose a Transformer-based state encoder as the backbone to connect the previous two components. It provides comprehensive state representations by modeling complicated relations between positive and negative feedback during the conversation.  Extensive experiments on three datasets demonstrate the advantage of our solution in serving new users, compared with a rich set of state-of-the-art CRS solutions. 

\end{abstract}





\author{Zhendong Chu}
\affiliation{%
  \institution{University of Virginia}
  \city{Charlottesville}
  \state{VA}
  \country{USA}
  }
\email{zc9uy@virginia.edu}

\author{Hongning Wang}
\affiliation{%
  \institution{University of Virginia}
  \city{Charlottesville}
  \state{VA}
  \country{USA}
  }
\email{hw5x@virginia.edu}

\author{Yun Xiao}
\affiliation{%
  \institution{JD.COM Silicon Valley R\&D Center}
  \city{Mountain View}
  \state{CA}
  \country{USA}
  }
\email{xiaoyun1@jd.com}

\author{Bo Long}
\affiliation{%
  \institution{JD.COM}
  \city{Beijing}
  \country{China}
  }
\email{bo.long@jd.com}

\author{Lingfei Wu}
\affiliation{%
  \institution{JD.COM Silicon Valley R\&D Center}
  \city{Mountain View}
  \state{CA}
  \country{USA}
  }
\email{lwu@email.wm.edu}

\maketitle

\input{intro}
\input{related_works}

\input{preliminary}

\input{method}

\input{experiments}

\section{Conclusion}
In this work, we present a meta reinforcement learning based solution to handle the problem of CRS policy learning in cold-start users. We learn a meta policy for generalization and fast adapt it on new users. We developed three components to ensure the efficiency and effectiveness of policy adaptation. First, a dedicated meta-exploration policy is adopted to identify the most informative user feedback for policy adaptation. Second, an adaptive state-aware recommendation component is built to quickly improve recommendation quality for new users. Third, a Transformer-based state encoder implicitly models positive and negative feedback collected in a conversation to precisely profile user preference. 

Currently our policy adaptation is performed independently across users; to further reduce its sample complexity, collaborative policy adaptation among users can be introduced to leverage observations among both new and existing users. As previous works reported improved CRS performance using a knowledge graph (KG), it is also interesting to study how personalized policy learning can benefit from KGs, e.g., adapts the entity relations in each user or design the meta-exploration strategy based on the KG.

\section{Acknowledgement}
We thank the anonymous reviewers for their insightful comments. This work was supported by NSF IIS-2007492 and NSF IIS-1838615.




\bibliographystyle{ACM-Reference-Format}
\bibliography{sample-base}


\end{document}

%% file: intro.tex
\section{Introduction}
\label{sec:intro}
While traditional recommendation solutions infer a user's preferences only based on her historically interacted items \cite{wu2020deja, sarwar2001item, rendle2010factorization, he2017neural, cai2021category, cai2018modeling, cai2020learning}, conversational recommender systems (CRS) leverage interactive conversations to adaptively profile a user's preference \cite{sun2018conversational, christakopoulou2016towards, lei2020estimation}. The conversations in CRS focus on questions about users' preferences on item attributes (e.g., brands or price range), in the form of pre-defined question templates \cite{sun2018conversational, lei2020estimation, deng2021unified} or timely synthesized natural language questions \cite{li2018towards, zhou2020towards}. Through a series of question answering, a profile about a user's intended item can be depicted, even when the user  is new to the system \cite{christakopoulou2016towards}, i.e., the cold-start users, which gives CRS an edge in providing improved recommendations. 

\citet{christakopoulou2016towards} first proposed the idea of CRS. Their solution focused on deciding what item to ask for feedback; and off-the-shelf metrics, such as upper confidence bound \cite{auer2002using}, were leveraged for the purpose. Following this line, reinforcement learning (RL) based methods become the mainstream solution recently for CRS. \citet{sun2018conversational}  built a policy network to decide whether to recommend an item, or otherwise which item attribute to ask about in each turn of a conversation. However, in these two early studies, the conversation is terminated once a recommendation is made, no matter whether the user accepts it or not. \citet{lei2020estimation} studied multi-round conversational recommendation, where CRS can ask a question or recommend an item multiple times before the user accepts the recommendation (considered as a successful conversation) or quits (considered as a failed conversation). This is also the setting of our work in this paper. To better address multi-round CRS, \citet{lei2020interactive} leveraged knowledge graphs to select more relevant attributes to ask across turns. \citet{xu2021adapting} extend \cite{lei2020estimation} by revising user embeddings dynamically based on users' feedback on attributes and items. And \citet{deng2021unified} unified the question selection module and the recommendation module in an RL-based CRS solution, which simplifies the training of CRS. However, all aforementioned RL-based methods rely on existing user embeddings to conduct conversations and recommendations, which are not applicable to new users.

\begin{figure}[!h]
    \centering
    \vspace{-1mm}
    \includegraphics[width=7.5cm]{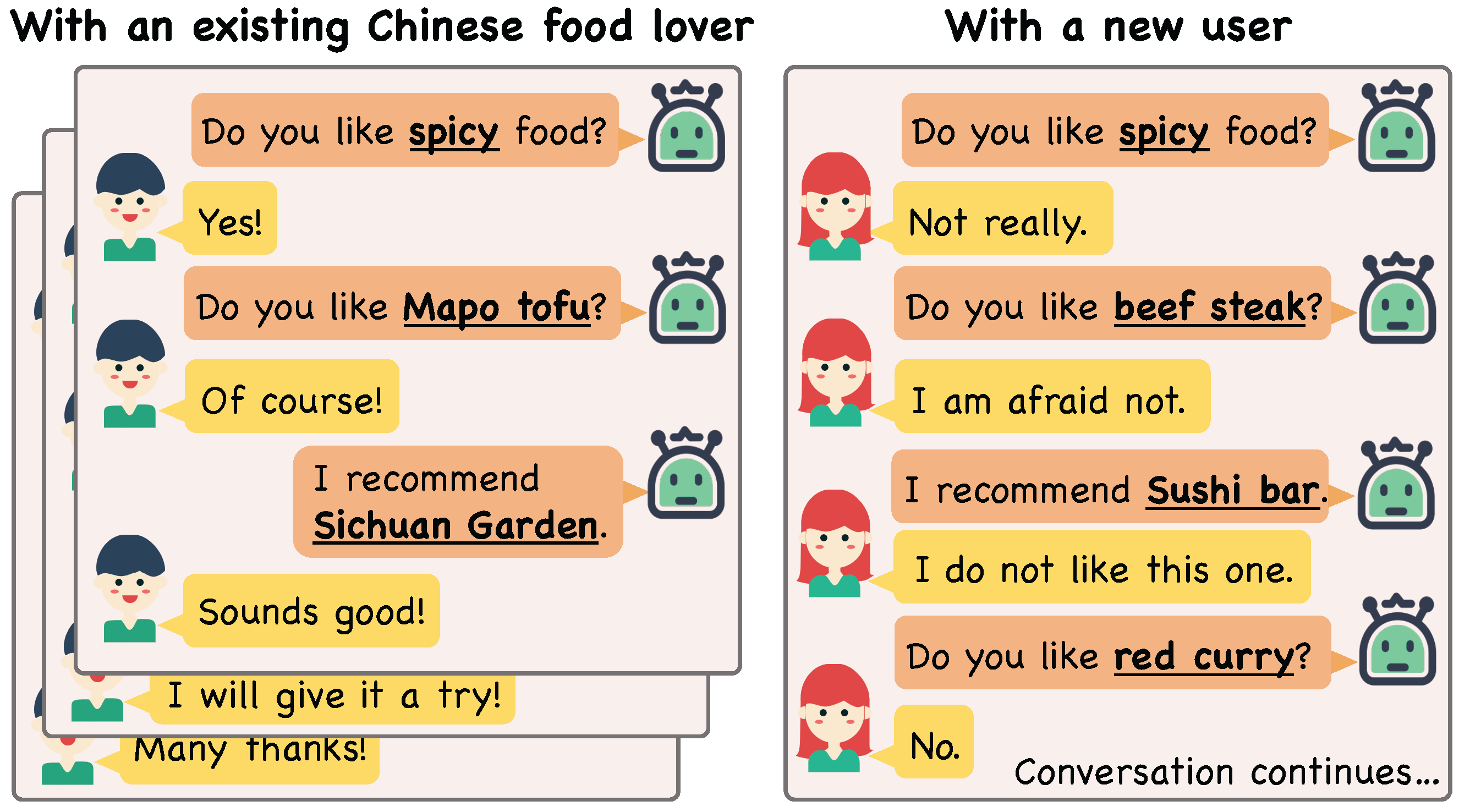}
    \vspace{-1mm}
    \caption{Example of cold-start CRS.}
    \label{fig:personalized_crs}
    \vspace{-3mm}
\end{figure}

Although CRS is expected to address the cold-start problem in recommendation, by profiling a new user via eliciting her preference about item attributes, how to acquire the most effective feedback to profile a single user still encounters the cold-start problem. More specifically, due to the heterogeneity of different users' preferences, the same policy can hardly be optimal in finding the sequence of interactions (asking questions or making recommendations) for all users, especially for those who do not contribute to the policy training.  Consider the example shown in Fig.\ref{fig:personalized_crs}, a policy trained with a population of Chinese food lovers cannot effectively serve new users who do not have any preferences on Chinese food. Once the interaction trajectory deviates from those often encountered during training, the effectiveness of the globally learnt CRS policy deteriorates, so does the quality of its recommendations.

We attribute this new challenge as cold-start policy learning in CRS, which is completely non-trivial but unfortunately ignored in most previous CRS studies. The goal is clear, i.e., adapt a CRS policy for each new user; but there are at least three main technical barriers blocking us from the goal. Firstly, \emph{how to efficiently adapt a policy to new users?} The tolerance of users about a prolonged conversation or bad recommendations is limited \cite{schnabel2018short, li2015toward, gilotte2018offline, chu2021improve, chu2021learning}, since all users wish to get high-quality recommendations with the least effort (e.g., shorter conversations) \cite{tetard2009lazy}. Hence, one cannot expect a large number of observations for CRS policy learning in a single user. 
Secondly, \emph{how to effectively explore user preferences for policy adaptation?} As shown in Fig.\ref{fig:personalized_crs}, successfully adapting a CRS policy to a new user depends on the user's preference, which however is elicited by the policy itself. This forms a \emph{chicken-and-egg} dilemma \cite{liu2021decoupling} and adds another layer of consideration when acquiring user feedback: before identifying what item the user is looking for, one first needs to figure out what policy best suits for the inquiry. 
Thirdly, \emph{how to decouple the adaptation of the conversation component and recommendation component in a CRS policy?} The conversation component (i.e., conversational policy) in CRS is  to profile a user by actively eliciting her feedback, while the recommendation component (i.e., item recommender) is to identify the most relevant recommendations based on the profile. Adaptation in both components is needed for new users, but the strategy for adapting them could be different for respective goals. 


In this paper, we address the problem of CRS policy learning for cold-start users via meta reinforcement learning \cite{liu2021decoupling,humplik2019meta,wang2016learning}, and name the proposed solution \textbf{\model{}}.
For the first challenge, we propose to learn a meta policy for CRS from a population of users and adapt it to new users with only a few trials of conversational recommendations.  The meta policy can be viewed as a starting point close to every single user's personalized policy. It thus builds the basis for efficient policy adaptation with only a handful of observations in each new user. 
Secondly, to acquire the most informative feedback for policy adaptation, we design a meta-exploration policy to identify user preferences via a few exploratory conversations. 
Thirdly, in addition to the CRS policy, we also adapt the recommendation module in each user to maximize the recommendation quality. 
To support such a decoupled adaptation strategy, 
we design a Transformer-based \cite{vaswani2017attention} state encoder as the backbone, which 
communicates the training signals between the conversation and recommendation components.


To evaluate the effectiveness of the proposed model, we compared \model{} with several state-of-the-art baselines for CRS on three public datasets. The results strongly demonstrated the advantage of our solution in making satisfactory recommendations to new users in CRS with a reduced number of conversations. We also conducted extensive ablation analysis on each proposed component to inspect its contribution on the improved performance: 1) the meta-exploration policy elicit informative user feedback for fast policy adaptation; and 2) the adapted recommendation component makes better recommendations by cooperating with the adapted conversation component.


%% file: related_works.tex
\section{Related works}
\noindent\textbf{Exploration-Exploitation (EE) Trade-off in CRS.} CRS take advantage of conversations with users to elicit their preferences in real time for improved recommendations. The main research effort in CRS focuses on addressing the explore-exploit trade-off in collecting user feedback. The first attempt made by \citet{christakopoulou2016towards} employed multi-armed bandit models to acquire users' feedback on individual items. A follow-up study \cite{zhang2020conversational} set an additional bandit model to select attributes to collect user feedback and employed a manually crafted function to decide when to ask questions or make recommendations. 
\citet{li2021seamlessly} unified attributes and items in the same arm space and let a bandit algorithm determine when to do what. Follow-up works \cite{zhao2022knowledge, wu2021clustering} also explored clustered and knowledge-aware conversational bandits.

\noindent\textbf{Meta learning for recommendation.} Meta learning \cite{finn2017model,chen2022robust} has been widely used to solve the cold-start problem in recommender systems. \citet{vartak2017meta} studied the item cold-start problem (i.e., how to recommend new items to users). They proposed two adaptation approaches. One learns a linear classifier whose weights are determined by the items represented to the user before and adapts the classifiers' weights for each user. Another one learns user-specific item representations and adapts the bias terms in a neural network recommender for the purpose. \citet{lee2019melu} separated the representation layer and decision-making layer in a neural recommendation model, and executed local adaptation on the decision-making layer for each new user.
\citet{zou2020neural} focused on interactive item recommendation, where the meta model is optimized by maximizing the cumulative rewards in each user. \citet{kim2022meta} deployed meta learning to online update recommender, where the meta learning rates are adaptively tuned on a per parameter and instance basis. To the best of our knowledge, we are the first to propose to tackle with cold-start CRS policy learning using meta reinforcement learning.

%% file: preliminary.tex
\section{Preliminary}
In this section, we first formulate the problem of multi-round CRS as a reinforcement learning problem, and then illustrate the concept of meta reinforcement learning and how we use it to address the cold-start challenge in CRS. 

\subsection{Problem Definition}
\label{sec-prob-def}
In this work, we study the problem of multi-round conversational recommendation \cite{lei2020estimation}, where CRS can ask questions or make recommendations multiple times before the user accepts the recommendation or quits the conversation. Similar to traditional recommender systems, CRS face a set of users $\mathcal{U}$ and a set of items $\mathcal{V}$; and we denote a specific user as $u$ and a specific item as $v$. Each item $v$ is associated with a set of pre-defined attributes $\mathcal{P}_v$. Attributes describe basic properties of items, such as movie genres in movie recommendations and authors in book recommendations. 

We formulate the CRS problem by a Markov decision process (MDP) \cite{deng2021unified, lei2020interactive, huai2020malicious, yao2021reversible}, which can be fully described by a tuple $(\mathcal{S}, \mathcal{A}, \mathcal{T}, \mathcal{R})$. $\mathcal{}S$ denotes the state space, which summarizes the conversation between the system and user so far. $\mathcal{A}$ denotes the action space for the system, which includes recommending a particular item or asking a specific attribute for feedback. $\mathcal{T}: \mathcal{S} \times \mathcal{A} \to \mathcal{S}$ is the state transition function, and $\mathcal{R}: \mathcal{S}\times \mathcal{A} \to [-R_{max}, R_{max}]$ is a bounded reward function suggesting a user's feedback on the system's actions. 
As we focus on meta policy learning for CRS in this work, how to best define reward is not our objective. We follow the reward function defined in \cite{lei2020estimation, lei2020interactive, deng2021unified}. In particular, we include the following rewards: (1) $r_{\text{rec\_suc}}$, a large positive reward when the recommended item is accepted; (2) $r_{\text{rec\_fail}}$, a negative reward when the recommended item is rejected; (3) $r_{\text{ask\_suc}}$, a positive reward when the inquired attribute is confirmed by the user; (4) $r_{\text{ask\_fail}}$, a negative reward when the inquired attribute is dismissed by the user; (5) $r_{\text{quit}}$, a large negative reward when the user quits the conversation without a successful recommendation.

With this formulation, a conversation in CRS can be represented as $d = \{(a_1, r_1), ... (a_T, r_T)\}$, where $T$ is the maximum number of  allowed turns. A conversation (or an episode in the language of RL, which we will use  exchangeablely) will terminate when (1) the user accepts the recommended item; or (2) the agent runs out of maximum allowed turns.  At each time step $t$, the CRS agent, which can be fully described by a policy $\pi(a_t|s_t)$, selects an action $a_t$ based on the current state $s_t$. The training objective of a CRS policy is to maximize the expected cumulative rewards over the set of observed episodes $D$, i.e.,
$$
    \mathcal{L}(\pi) = -\underset{d\sim P(D)}{\mathbb{E}} \big [ \sum_{t=0}^T R_t \big ],
    \label{eq:raw_loss}
$$
where $R_t=\sum_{t'=t}^T\gamma^{T-t'}r(a_t)$ is the accumulated reward from turn $t$ to the final turn $T$, and $\gamma\in[0, 1]$ is a discount factor to emphasize rewards collected in a near term. 


\subsection{Meta Reinforcement Learning for CRS}
\label{sec:usp_meta}
Instead of learning a single global policy $\pi$, we propose to learn personalized policy $\pi_u$ for each user $u$ (new or existing) to address the cold-start challenge for CRS. The fundamental reason that almost all previous works \cite{lei2020estimation, lei2020interactive, deng2021unified} focused on global policy learning is that they (implicitly) assumed users know all attributes of their desired items and share the same responses over those attributes; in other words, user feedback is fully determined by the item. 
This assumption is unrealistically strong and naive, since different users can describe the same item very differently, because of their distinct knowledge and preferences. For example, some users choose a mobile phone for its appearance while others choose it because of its brand. As a result, a global policy can hardly be optimal for every single user, especially the new users whose preferences are not observed during global policy training.  
In this work, we impose a weaker and more realistic assumption about users' decision making by allowing user-specific feedback $\mathcal{R}_{\mathcal{U}}$, which calls for personalized policies. Therefore, a personalized policy for user $u$ should minimize,
\begin{equation}
    \mathcal{L}_u(\pi) = -\underset{d\sim P(\mathcal{D}_u)}{\mathbb{E}} \big [ \sum_{t=0}^T R_u(a_t) \big ],
    \label{eq:usp_loss}
\end{equation}
where $\mathcal{D}_u$ is a collection of conversations from user $u$ and $R_u(a_t)=\sum_{t'=t}^T\gamma^{T-t'}r_u(a_{t'})$. 
To find the best personalized policy $\pi_u$ (parameterized by $\theta_u$) for each new user $u$, instead of learning from scratch every time, we choose to learn a meta policy parameterized by $\theta$ and use it as a starting point to look for $\theta_u$. Following the convention of meta learning, we assume a set of conversations $\mathcal{D}_u^s$ (i.e., the support set) for policy adaptation , in addition to the set $\mathcal{D}_u^q$ (i.e., the query set) for policy evaluation. Hence, the size of support set $\mathcal{D}_u^s$ in each user $u$ denotes the conversation budget for us to find $\theta_u$ when serving a single user. Given limited tolerance of an ordinary user to prolonged conversations, a performing solution should find the optimal $\theta_u$ with the size of $\mathcal{D}_u^s$ \emph{as small as possible}. In the meta-train phase, we conduct local adaptation from the meta policy on the $\mathcal{D}_u^s$ of the existing users (i.e., training users), and then evaluate and update the meta policy on training users' $\mathcal{D}_u^q$. In the meta-test phase, we test the meta policy on the new users (i.e., testing users) by executing local adaptation on their support sets, and then test the obtained local policy on their query sets.




%% file: method.tex
\begin{figure}[!htp]
    \centering
    \includegraphics[width=8.5cm]{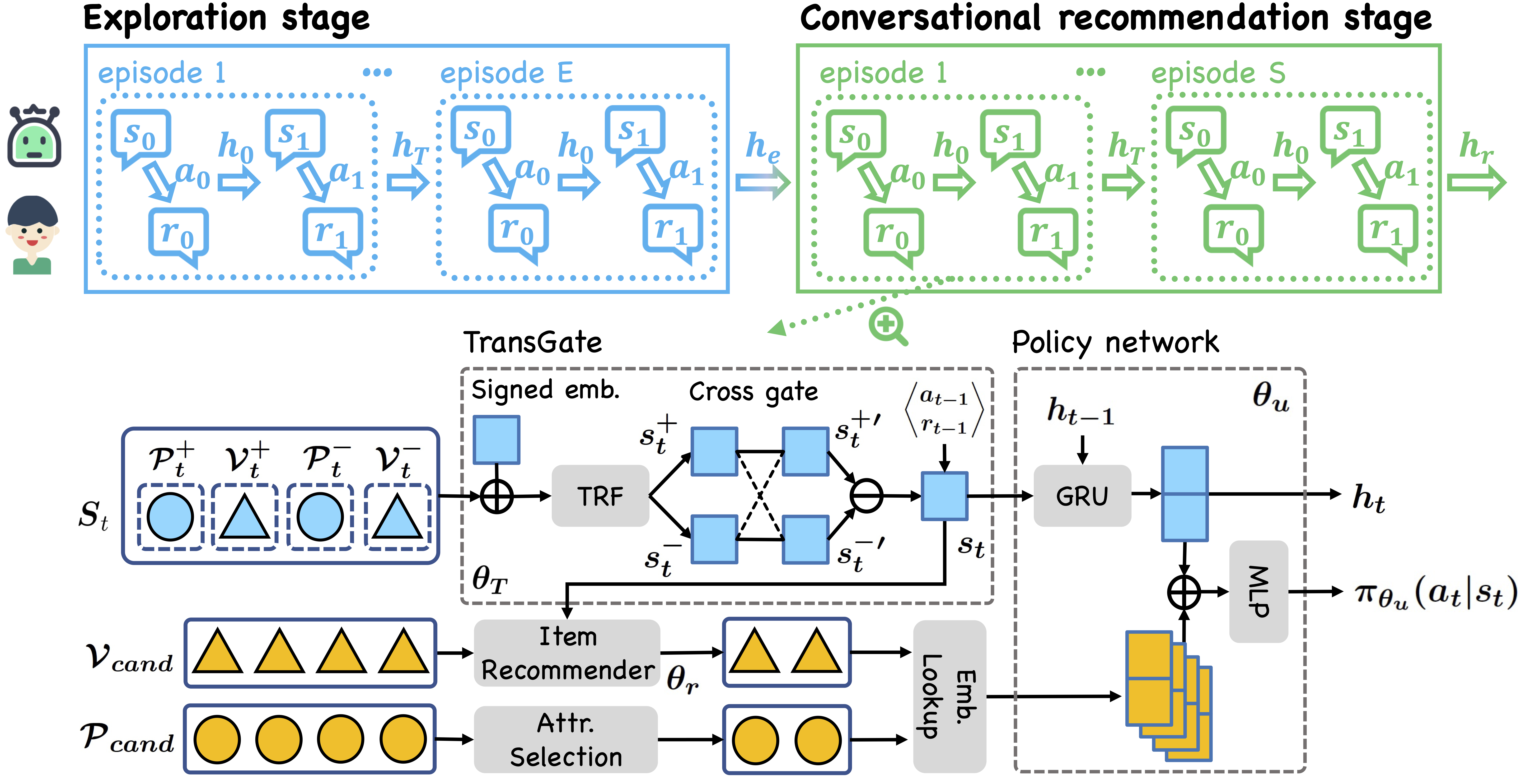}
    \caption{The workflow of \model{} training. Each user's support set is separated into the exploration stage and conversational recommendation stage. The last hidden state from the previous episode is passed to the next episode as its initial state throughout the course of \model{} in each user. }
    \label{fig:overview}
\end{figure}

\section{Methodology}
In this section, we describe the design of \model{} in detail. We first introduce our two-stage meta policy learning framework designed for cold-start CRS. Then, we describe the details of the state-based item recommender, which is separately adapted to maximize the recommendation quality. The Transformer-based state encoder, which aims to rapidly capture a user's preference from her both positive and negative feedback in a conversation, is lastly explained. Fig.\ref{fig:overview} shows the overview of MetaCRS.



\subsection{Two-stage Meta Policy Learning for CRS}
\label{sec:meta}
Motivated by the seminal work Model-Agnostic Meta-Learning (MAML) \cite{finn2017model}, we propose to first learn a meta CRS policy from a population of users; and then for each individual user, we adapt the meta policy to a user-specific policy  with only a few trails of conversational recommendations with the user. We obtain the meta policy by maximizing the policy adaptation performance in a given set of training users. Specifically, in each user $u$, we perform policy adaptation on her support set $\mathcal{D}_u^s$, where $\theta_u$ is initialized with $\theta$ and then updated by optimizing Eq.\eqref{eq:usp_loss} via gradient descent. 
The gradient for Eq.\eqref{eq:usp_loss} is computed by the REINFORCE \cite{williams1992simple} algorithm,
\begin{equation}
    \nabla_{\theta_u}\mathcal{L}_u(\pi_{\theta_u}) = -\mathbb{E}\Big [ \sum_{t=0}^TR_u(a_t)\nabla_{\theta_u}\log \pi_{\theta_u}(a_t|s_t)\Big ].
    \label{eq:gradient}
\end{equation}
The gradient of the meta policy with respect to $\theta$ (i.e.,  $\nabla_{\theta}\mathcal{L}_u(\theta_u)$) is computed in the same way as Eq.\eqref{eq:gradient}, but on the corresponding query set $\mathcal{D}_u^q$. In this way, the meta policy is optimized for generalization, akin to cross-validation. Note that to exactly compute the gradient for $\theta$, we need to take a higher-order derivative in $\nabla_{\theta_u}\mathcal{L}_u(\theta_u)$ with respect to $\theta$ on the support set as well, since $\theta_u$ is a function of $\theta$. In this work, we followed the the first-order approximation methods proposed in \cite{finn2017model,nichol2018first} to simplify the gradient computation. 

In meta-learning for supervised learning tasks, e.g., image classification \cite{finn2017model,nichol2018first,vuorio2019multimodal}, the support set and query set are predefined and thus not affected by the learnt models. Therefore, gradient-based optimization alone is sufficient for meta model learning and adaptation. But in our problem, what we will observe in $\mathcal{D}_u^s$ and $\mathcal{D}_u^q$ are completely determined by the employed policy $\pi_{\theta_u}$, which however is supposed to be derived from $\mathcal{D}_u^s$ and $\mathcal{D}_u^q$. This causes the so-called \emph{chicken-and-egg} dilemma \cite{liu2021decoupling} for meta policy learning which we discussed in the introduction, and calls for additional treatments beyond gradient-based policy optimization. 

Potential bias in the currently learnt policy prevents it from being effective in acquiring the most informative feedback for meta policy learning and adaptation. Hence, we propose to separate  policy adaptation in each user into an \emph{exploration} stage and a \emph{conversational recommendation} stage, and design their corresponding policies. To avoid ambiguity, we refer to the policy for the exploration stage as the meta-exploration policy (denoted as $\pi_{\theta_e}$), and the policy for the conversational recommendation stage as the CRS policy (denoted as $\pi_{\theta_u}$). 
This is similar to the explore-then-commit strategy \cite{garivier2016explore,lattimore2020bandit} in bandit literature. But note that our meta-exploration policy is \emph{not} personalized, as its sole goal is to quickly identify what kind of user the system is interacting with. Hence, we choose to estimate it from the whole set of training users. 
In particular, we reserve the first few episodes in each user's support set for our exploration stage, denoted as the exploration set $\mathcal{D}_u^e$. The size of $\mathcal{D}_u^e$ is a hyper-parameter to be tuned for different CRS applications. $\mathcal{D}_u^e$ will only be used to estimate the meta-exploration policy $\pi_{\theta_e}$.

In \model{}, the meta-exploration policy $\pi_{\theta_e}$, meta CRS policy $\pi_{\theta}$ and personalized CRS policies $\{\pi_{\theta_u}\}_{u\in\mathcal{U}}$ are realized by the same RNN-based policy network architecture \cite{hopfield1982neural, duan2016rl, wang2016learning} with Gated Recurrent Units (GRUs) \cite{bahdanau2014neural, duan2016rl} to better encode the conversation history, but we estimate different parameters for them respectively. To avoid ambiguity, we will use the learning of meta-exploration policy $\pi_{\theta_e}$ as an illustrating example; and the same procedure applies to the learning of other policies.

Specifically, at the $t$-th turn of an episode, we observe a new state and encode it using a state encoder. We leave the discussion about our state encoding in Section \ref{sec:encoder}. 
The encoded state $s_t$ is 
provided as input to the policy network. 
The output $h_t$ of the GRU is fed to a fully connected layer followed by a softmax function to produce the action distribution for $\pi_{\theta_e}(a_t|s_t)$. 
In each user, by the end of each episode, the GRU's last output hidden state $h_T$ \footnote{If the conversation ends before the maximum turn, $h_T$ stands for the latent state at the successful recommendation.} is passed to the user's next episode as its initial state, such that this user's conversation history with the system can be continuously used to jump start her next conversation. 
Enabled by this design, information collected from the exploration stage is passed over to the conversational recommendation stage to profile who the user is (denoted as $h_e$), and then to the query set (denoted as $h_r$) to suggest what the user's preference could be. This sequential process is depicted in Fig.\ref{fig:overview}.


The policy networks for $\pi_{\theta}$ and  $\{\pi_{\theta_u}\}_{u\in\mathcal{U}}$ are trained via the meta learning procedure described at the beginning of this section, on top of the rewards defined in Section \ref{sec-prob-def}. But the meta-exploration policy $\pi_{\theta_e}$ is trained with a specially designed reward function, as its sole purpose is to identify what kind of user the system is serving. 
Inspired by \cite{liu2021decoupling, kamienny2020learning, humplik2019meta}, we adopt pre-trained user embeddings $\{e_u\}_{u\in\mathcal{U}}$ obtained on users with historical observations (i.e., training users) to design the exploration reward,
\begin{equation}
\label{eq-exp-reward}
    r_e(s_t) = \log{P(e_u|s_t)}- \log{P(e_u|s_{t-1})}, 
\end{equation}
where 
$
    P(e_u|s_t) = \frac{\text{exp}({h_t^\top e_u})}{\sum_{u\in\mathcal{U}}\text{exp}({h_t^\top e_u})}
$. Note here we use the GRU's output hidden state $h_t$ to predict the user embedding, just as how we use it to construct the policies.
Specifically, we obtain $\{e_u\}_{u\in\mathcal{U}}$ from a Factorization Machine model  \cite{lei2020estimation} trained on observed user-item interactions in training users. More details about this user embedding learning can be found in Section \ref{sec:data}. The insight behind our exploration reward design is that we promote the actions that help us identify a specific user during the exploration stage. 
Following the suggestion from \cite{liu2021decoupling, kamienny2020learning, humplik2019meta}, we also add a cross entropy loss on the meta-exploration policy network's latent state $h_t$ to regularize the estimation of $\theta_e$,
\begin{equation}
    \mathcal{L}_e(\pi_{\theta_e}) = -\mathbb{E}\Big[\sum_{t=0}^TR_e(s_t) + \sum_{t=0}^T\log P(e_u|h_t) \Big ],
    \label{eq:exp_loss}
\end{equation}
where $R_e(s_t)$ is the accumulated discounted reward based on Eq.\eqref{eq-exp-reward} from turn $t$. The gradient of the first term is also computed by the REINFORCE algorithm.

\subsection{State-aware Item Recommender}
Previous studies use a pre-trained recommender through the course of CRS \cite{lei2020estimation, lei2020interactive, xu2021adapting}, as their focus is mostly on deciding when to make a recommendation or otherwise what question to ask, i.e., the conversation component. A pre-trained recommender restricts the CRS policy to accommodate the recommender's behavior, which adds unnecessary complexity for policy adaptation. Such a black box design slows down personalized policy learning. For example, a user wants a phone of a specific brand, but the recommender regards \emph{brand} as an unimportant attribute. It is difficult for CRS to recommend successfully even though the policy already elicits her preferences, which will in turn hurts the policy adaptation since the episode is failed. Hence, it is crucial to also local adapt the recommender to learn to make high-quality recommendations in cooperation with the conversation component. 

In \model{}, we set a learnable item recommender to rank candidate items based on the state embedding from the state encoder, which will be explained in Section \ref{sec:encoder}. The ranking score of an item $v$ is calculated by,
\begin{equation*}
    w_t(v) = e_v^\top(\boldsymbol{W}_1 s_t + \boldsymbol{b}_1),
\end{equation*}
where $\{\boldsymbol{W}_1, \boldsymbol{b}_1\}$ are learnable parameters for the recommender, collectively denoted as $\theta_r$; $s_t$ is the state embedding obtained from the state encoder. 
We perform local adaptation on $\theta_r$ to obtain a personalized recommender, by minimizing the following cross-entropy loss once a successful conversation concludes,
\begin{equation}
    \mathcal{L}_r(\theta_r) = -
    \frac{1}{T_{s}}\mathbb{I}(T_s \leq T)\sum_{t=0}^{T_{s}}\log \frac{\text{exp}(w_t(v_s))}{\sum_{|\mathcal{V}^+_t|}\text{exp}(w_t(v))},
    \label{eq:ranker_loss}
\end{equation}
where $T_{s}$ is the index of the successful turn and $v_{s}$ is the accepted item. This loss function encourages the adapted recommendation component to identify the finally accepted item as early as possible in a conversation.
We denote the meta parameters of $\theta_r$ as $\theta_R$.

\subsection{TransGate State Encoder}
\label{sec:encoder}
Previous solutions \cite{zhao2018recommendations, zhang2021multi, xu2021adapting} have shown the power of negative feedback in CRS state modeling. It is even more important for cold-start CRS, especially in the early stage of  policy adaptation when the policy is more likely to collect negative feedback. Ineffective modeling of negative feedback will slow down policy adaption. Moreover, positive and negative feedback posits distinct information about users' preference, and thus calls for different treatments. 
We employ a Transformer to model such complicated relations in an ongoing conversation into a state, with a cross gate mechanism to differentiate the impact from positive and negative feedback. We name this state encoder as TransGate.

At turn $t$, we accumulate four kinds of feedback from a user in this conversation: (1) $\mathcal{P}^+_{t}$, attributes confirmed by the user; (2) $\mathcal{V}^+_t$, candidate items satisfying all accepted attributes; (3) $\mathcal{P}_t^-$, attributes dismissed by the user; (4) $\mathcal{V}_t^-$, items rejected by the user.
Collectively, we denote $\mathcal{S}_t = \{\mathcal{P}^+_{t}, \mathcal{V}^+_t, \mathcal{P}_{t}^-, \mathcal{V}_{t}^-\}$. 
We first map elements in $\mathcal{S}_t$ into vectors $e$ with an embedding layer, where attribute and item embeddings are pre-trained with training users' historical observations. Candidate items and rejected items are aggregated separately to reduce the sequence length, 
\begin{equation*}
    e_{\mathcal{V}}^{+} = \frac{1}{|\mathcal{V}^{+}_t|}\sum_{v\in \mathcal{V}^{+}_t}e_{v}^{+}, \;
    e_{\mathcal{V}}^{-} = \frac{1}{|\mathcal{V}^{-}_t|}\sum_{v\in \mathcal{V}^{-}_t}e_{v}^{-}
    .
\end{equation*}
In the original Transformer \cite{vaswani2017attention}, elements are encoded with position embeddings. In our case, the order among the elements is not important, but encoding the sign of user feedback (i.e., accepted or rejected) is critical. Inspired by position embeddings, we propose to encode user feedback into signed embeddings $\{e^+, e^-\}$. We add $e^+$ to positive elements and $e^-$ to negative elements in $\mathcal{S}_t$. We use the current candidate items to provide positive context. Then, we feed the obtained embeddings into $L$ Transformer layers. For simplicity, we keep the notations of transformed embeddings unchanged. We then aggregate the positive and negative elements separately to obtain an embedding for positive feedback  and an embedding for negative feedback,
\begin{equation*}
    s_t^+ = \frac{1}{1+|\mathcal{P}^+_t|}\Big( e_{\mathcal{V}}^+ +\sum_{p\in \mathcal{P}^+_t} e_{p}^+\Big), \; s_t^- = \frac{1}{1+|\mathcal{P}^-_t|}\Big(e_{\mathcal{V}}^-+\sum_{p\in \mathcal{P}^-_t}e_{p}^-\Big).
\end{equation*}

The positive and negative feedback embeddings may contain overlapped information, which will confuse policy learning. For example, an item that already satisfies all confirmed attributes so far can still be rejected by the user. We propose a cross gate mechanism to further differentiate the positive and negative information
$
    s_{t}^{+\prime} = s_t^+ \odot g^-, ~~s_{t}^{-\prime} = s_t^- \odot g^+,
$
where $\odot$ denotes the element-wise product and $\{g^+, g^-\}$ are defined as 
$$g^+ = \sigma(\boldsymbol{W}_2 s_t^+ + \boldsymbol{b}_2), ~~ g^- = \sigma(\boldsymbol{W}_3 s_t^- + \boldsymbol{b}_3),$$ 
where $\sigma(\cdot)$ is the sigmoid function and $\{\boldsymbol{W}_2, \boldsymbol{W}_3, \boldsymbol{b}_2, \boldsymbol{b}_3\}$ are learnable parameters. We obtain the final state by
$
    s_t = s^{+\prime}_t - s^{-\prime}_t.
$ 
The set of parameters for the TransGate encoder is denoted as $\theta_T$, which is learnt from the conversations with training users. We should note once learnt this encoder is shared globally by all users without personalization. The state embedding is then concatenated with the encoding of $\left<a_{t-1}, r_{t-1}\right>$ as the input to the RNN-based policy network. In particular, the action embedding is directly read off based on the pre-trained attribute and item embeddings, and we set a linear layer to encode the reward.

\begin{algorithm}[!hpt]
\SetAlgoLined
\Require User population $\mathcal{U}$, learning rates $\alpha, \beta$, meta parameters $\theta$, $\theta_e$, $\theta_R$, $\theta_T$\;
 \While{not \textup{Done}}{
    Sample a batch of users $\mathcal{U}_b\sim P(\mathcal{U})$\;
  \For{\textup{each} $u \in \mathcal{U}_b$}{
    Collect $\mathcal{D}_u^e$ and $h_e$ by executing $\pi_{\theta_e}$ \;
    Initialize $\theta_u=\theta$, $\theta_r=\theta_R$\;
    Collect $\mathcal{D}_u^s$ and $h_r$ by executing $\pi_{\theta_u}$ with $h_e$\;
    Evaluate $\nabla_{\theta_u}\mathcal{L}_u$ and $\nabla_{\theta_r}\mathcal{L}_r$ using $\mathcal{D}_u^s$\;
    Compute adapted parameters with gradient descent: $\theta_u = \theta_u- \alpha\nabla_{\theta_u}\mathcal{L}_u$, $\theta_r = \theta_r- \alpha\nabla_{\theta_r}\mathcal{L}_r$                                           \;
    Collect $\mathcal{D}_u^q$ by executing $\pi_{\theta_u}$ with $h_r$\;
  }
  Update $\theta, \theta_R, \theta_T$ using each $\mathcal{D}_u^q$ by minimizing $\mathcal{L}_u$,  $\mathcal{L}_r$\;
  Update $\theta_e, \theta_T$ using each $\mathcal{D}_u^e$ by minimizing $\mathcal{L}_e$\;
 }
 \caption{Optimization algorithm of MetaCRS}
 \label{alg:metacrs}
\end{algorithm}

\subsection{Optimization Algorithm}


Now we are finally equipped to illustrate the complete learning solution for MetaCRS in Algorithm \ref{alg:metacrs}. In the inner for-loop, we perform policy adaption to obtain the personalized CRS policy (including item recommender). In the outer while-loop, we update all meta parameters. To simplify the gradient computation, we stop the gradients on the inherited initial hidden state $h_T$ from the latest episode in back-propagation.
In practice, we update the local parameters once an episode is executed, as we find empirically it works better than updating once after  the whole $\mathcal{D}_u^s$ is finished. When serving new users in the meta-test phase, we fix $\{\theta_e, \theta_T\}$ and only execute local adaptation (the inner for-loop part in Algorithm \ref{alg:metacrs}) with the corresponding parameters initialized by $\{\theta, \theta_R\}$. 

In each turn, we use all the candidate items $\mathcal{V}_{cand}$ (i.e., $\mathcal{V}_t^+$) and attributes $\mathcal{P}_{cand}$ to construct the action space, where $\mathcal{P}_{cand}$ is the entire attribute set excluding $\mathcal{P}_t^+$ and $\mathcal{P}_t^-$. \citet{deng2021unified} reported that a very large action space always slowed down policy learning. To generate a reasonable action space, we follow the manually crafted rules from \cite{deng2021unified} to select $K_{A}$ attributes from $\mathcal{P}_{cand}$ and select the top-$K_{I}$ items provided by the state-based item recommender.

%% file: experiments.tex
\section{Experiments}
To fully demonstrate the effectiveness of \model{} in solving the cold-start CRS problem, we conduct extensive experiments and study the following four research questions (RQ):

\begin{itemize}[noitemsep,leftmargin=*,topsep=0pt]
    \item \textbf{RQ1}: Can MetaCRS achieve better performance than state-of-the-art CRS solutions when handling new users? 
    \item \textbf{RQ2}: Does our meta reinforcement learning based adaptation strategy work better than other adaptation strategies?
    \item \textbf{RQ3}: How quickly can \model{} obtain a good personalized policy for each user?
    \item \textbf{RQ4}: How does each proposed component contribute to the final performance of \model{}?
\end{itemize}

\begin{table}[!htp]
    \vspace{-2mm}
    \caption{Summary statistics of datasets.}
    \vspace{-2mm}
    \centering
    \begin{tabular}{l@{\hspace{1\tabcolsep}} r@{\hspace{1\tabcolsep}} r@{\hspace{1\tabcolsep}} r@{\hspace{1\tabcolsep}}}
    \toprule
     & \textbf{LastFM}   & \textbf{BookRec} &\textbf{MovieLens} \\ \midrule
    \#Users  & 1,801  & 1,891 & 3,000  \\
    \#Items  &  7,432  &  4,343 & 5,974 \\
    \#Attributes & 33 & 35 & 35  \\ 
    \#Interactions & 72,040 & 75,640 & 120,000 \\ \midrule
    Avg. $|\mathcal{P}_u|$ & 7 & 8 & 12 \\
    Avg. $|\mathcal{P}_v|$ & 4.07 & 8.15 & 5.02 \\
     Avg. $|\mathcal{P}_o|$ & 5.44 & 5.30 & 4.25 \\
    \bottomrule
    \end{tabular}
    \label{tab:datasets}
    \vspace{-2mm}
\end{table}

\subsection{Datasets}
\label{sec:data}
We evaluate \model{} on three multi-round conversational recommendation benchmark datasets \cite{lei2020estimation, lei2020interactive, deng2021unified, zhang2021multi} and summarize their statistics in Table \ref{tab:datasets}. 
\begin{itemize}[noitemsep,leftmargin=*,topsep=0pt]
    \item \textbf{LastFM} \cite{Bertin-Mahieux2011} is for music recommendation. \citet{lei2020estimation} manually grouped the original attributes into 33 coarse-grained attributes. 
    \item \textbf{BookRec} \cite{he2016ups} is for book recommendation. We further processed it by selecting top 35 attributes according to their TF-IDF scores across items and filter out items with too few attributes.
    \item \textbf{MovieLens} \cite{harper2015movielens} is for movie recommendation. We performed the same pre-processing as on the BookRec dataset.
\end{itemize}

We randomly split users for training, validation and testing with the ratio 8:1:1, such that the evaluation set only contains new users. On each benchmark dataset, we obtained user, item and attribute embeddings (denoted as $e_{\mathcal{U}}, e_{\mathcal{V}}, e_{\mathcal{P}}$) using a variant of Factorization Machine (FM) proposed in \cite{lei2020estimation} on observed user-item interactions in the training set. Similar to \cite{lei2020estimation, lei2020interactive, deng2021unified}, we developed a user-simulator to generate conversations based on the observed user-item interactions in the dataset. We describe the simulator in detail in Section \ref{sec:simulator}. However, the number of observed interactions in each user is not even, which may cause the learned meta policy biased toward users with more observed interactions.
To better study the problem of personalized CRS policy learning, as part of our simulation, we generated 40 user-item interactions for each user by sampling items proportional to the score $e_u^\top e_v$ to augment the interaction data for our evaluation purpose. We should note such simulation design will not ease the necessity of personalized CRS policy learning, since $e_u$ is never directly disclosed to the policy. 
We provide our code and generated data to facilitate follow-up research and ensure the producibility of our reported results \footnote{\url{https://github.com/zdchu/MetaCRS.git}}.


\begin{table*}[t]
\vspace{-2mm}
\caption{Comparison of CRS performance among models on three datasets. * stands for the best performance in each group.}
\label{tab:overall}
\vspace{-2mm}
\begin{tabular}{@{}cc|cccc|cccc|cccc|c@{}}
\toprule
    &   &   MaxE & EAR & SCPR & UNI & ConUCB & ConTS & FPAN & UR & F-FT  & F-IA & UR-FT & UR-IA & MetaCRS \\ \midrule
\multirow{2}{*}{LastFM}  
& SR@10 & 0.137 & 0.428 &  0.432 & 0.441$^*$  & 0.237  & 0.270 &  0.508 &  0.641$^*$ & 0.533 & 0.529 &  0.613 & 0.678$^*$ &  \textbf{0.713}   \\
& AT    & 9.71 & 8.62 &  8.70 & 8.52$^*$   &  8.69  & 8.93 &  8.08   &  7.02$^*$  & 8.01 & 8.13 & 7.19 & 6.85$^*$ &  \textbf{6.18}  \\ \midrule
\multirow{2}{*}{BookRec}   
& SR@10 & 0.206 & 0.320 & 0.329 &  0.358$^*$  & 0.181  & 0.243 & 0.397$^*$   &   0.384  &  0.405 & 0.411 &  0.417 & 0.420$^*$ &  \textbf{0.487} \\
& AT    & 9.64  &    9.01 & 9.11 & 9.00$^*$ & 9.52   & 9.17 & 8.31$^*$  &  8.55  & 8.36$^*$ & 8.52 & 8.54 & 8.41 & \textbf{8.06} \\ \midrule
\multirow{2}{*}{MovieLens} 
& SR@10 &  0.262 &  0.552 & 0.545  &   0.596$^*$   & 0.272 & 0.434  &  0.589 &  0.681$^*$  & 0.603 & 0.612 & 0.677 & 0.704$^*$ &  \textbf{0.745}  \\
& AT  & 9.46 & 7.98 & 7.89$^*$  & 8.01  &  8.36 & 8.08 & 7.81   &    7.00$^*$  & 7.78  & 7.69 & 7.14 & 6.88$^*$ &    \textbf{6.27}   \\ \bottomrule
\end{tabular}
\end{table*}

\subsection{Experiment Settings}
\subsubsection{User simulator}
\label{sec:simulator}
CRS needs to be trained and evaluated via interactions with users. 
Previous simulator designs are item-centric \cite{lei2020estimation, lei2020interactive, deng2021unified}, enforcing all users to respond in the same way to all attributes of target item $v$ (i.e., confirming every entry in $\mathcal{P}_v$). This setting is unrealistically restrictive and eliminates the necessity of personalized policies. 
To demonstrate the utility of personalized CRS policy learning, we design a \emph{user-centric} simulator that supports user-specific feedback in each conversation. 

In detail, we used the pre-trained user and attribute embeddings to generate each user's preferred attribute set $\{\mathcal{P}_u\}_{u\in\mathcal{U}}$, by selecting the top-ranked attributes for each user based on the score $e_u^{\top}e_p$. During the course of CRS, the simulated user will only confirm the overlapped attributes in $\mathcal{P}_o =  \mathcal{P}_u \cap \mathcal{P}_v$, and dismiss all others. On the BookRec dataset, because the original entries in $\mathcal{P}_v$ is too generic to be informative, i.e., too many attributes appear in almost all items, we decided to also increase $\mathcal{P}_v$ on this dataset by adding top-ranked attributes for each item based on the score $e_v^{\top}e_p$. We report the mean value of $|\mathcal{P}_u|$, $|\mathcal{P}_v|$ and $|\mathcal{P}_o|$ resulted from our simulation  on each dataset in Table \ref{tab:datasets}. 




\subsubsection{Baselines} To fully evaluate the effectiveness of \model{}, we compared it with a set of representative baselines. We categorized the baselines into three groups for different comparison purposes. In the first group, we compared MetaCRS with a rich set of state-of-the-art CRS methods to answer RQ1: 
\begin{itemize}[noitemsep,leftmargin=*,topsep=0pt]
    \item \textbf{Max Entropy (MaxE)} is a rule-based method suggested in \cite{lei2020estimation}. In each turn, the attribute with maximum entropy is to be asked or top-ranked items are to be recommended based on the rule.
    \item \textbf{EAR} \cite{lei2020estimation} is a three-stage solution consisting estimation, action and reflection steps. It updates the conversation and recommendation components  using reinforcement learning.
    \item \textbf{SCPR} \cite{lei2020interactive} reformulates the CRS problem as an interactive path reasoning problem on the user-item-attribute graph. Candidate attributes and items are selected according to their relations with collected user feedback on the graph.
    \item \textbf{UNICORN (UNI)} \cite{deng2021unified} integrates the conversation and recommendation components into a unified RL agent. Two heuristics for pre-selecting attributes and items in each turn are proposed to simplify its RL training.
\end{itemize}

Baselines in this group rely on pre-trained user embeddings to make recommendations or compute states, which are not available in new users. To apply them to new users, we used the average embedding of all training users as the embedding for new users. This group of baselines are learnt on training users and then evaluated on the testing users.

In the second group, we compared MetaCRS with solutions which handle new users by updating user embeddings dynamically within a conversation, such that they can provide adaptive recommendations. We consider the following algorithms:
\begin{itemize}[noitemsep,leftmargin=*,topsep=0pt]
    \item \textbf{ConUCB} \cite{zhang2020conversational} introduces the concept of super arms (i.e., attributes) to traditional bandit algorithms. Items and attributes with the highest upper confidence bound are selected. The attributes are asked in a fixed frequency by a hand-crafted function.
    \item \textbf{ConTS} \cite{li2021seamlessly} overcomes ConUCB's limitation by replacing the hand-crafted function with a Thompson sampling procedure. The user embeddings of cold-start users are updated with users' feedback on the asked attributes and items.
    \item \textbf{FPAN} \cite{xu2021adapting} extends the EAR model by utilizing a user-item-attribute graph to enhance offline representation learning. User embeddings are revised dynamically based on users' feedback on items and attributes in the conversation.
\end{itemize}
Our TransGate and state-based item recommender can also dynamically capture user preferences, and provide adaptive recommendations within a conversation. To further study their value in learning personalized CRS policies, we integrated them with UNICORN, and denoted this variant as \textbf{UR}, which is also included in the second group. All baselines in this group used the same pre-trained embeddings as MetaCRS. To improve the practical performance of ConUCB and ConTS, we adopted the heuristics in \cite{deng2021unified} to pre-select arms according to the similarity with accepted attributes. 

To answer RQ2, we equip FPAN and UR with the ability to adapt policies on new users via the following two widely used strategies, which forms the third group of baselines:

\begin{itemize}[noitemsep,leftmargin=*,topsep=0pt]
    \item \textbf{Fine-tuning (FT)}: We first pre-train a global policy on all training users. During testing,  we fine-tune the policy on the whole support set of all new users. 
    \item \textbf{Independent adaptation (IA)}: We first pre-train a global policy on all training users. For each new user, we perform continual training on her support set to obtain a personalized policy.
\end{itemize}

We denoted the resulted variants as \textbf{F-FT}, \textbf{F-IA} and \textbf{UR-FT}, \textbf{UR-IA} respectively. As we found policy gradient was more effective and efficient than UNICORN's original $Q$-learning based algorithm in our experiments,  we applied policy gradient for model update in all UNICORN-based baselines.



\subsubsection{Evaluation metrics} We followed the widely-used metrics in previous works \cite{lei2020estimation, lei2020interactive, deng2021unified} to evaluate the CRS solutions. We evaluated the average ratio of successful episodes within $T$ turns by success rate (SR@$T$). 
We also evaluated average turns in episodes (AT). A better policy is expected to recommend successfully with less turns. The length of failed conversations is counted as $T$.

\subsubsection{Implementation details} We performed the training of meta policy on training users, and local adaptation on validation and testing users. We selected the best model according to its validation performance. The query sets of testing users are used to obtain the final performance for comparison. We set the rewards as: $r_{\text{rec\_suc}}=1$, $r_{\text{rec\_fail}}=-0.1$, $r_{\text{ask\_suc}}=0.1$, $r_{\text{rec\_fail}}=-0.1$, $r_{\text{quit}}=-0.3$. The action embedding
size and the hidden size are set to be 64 and 100, while reward embedding size is set to 10. We set 1 Transformer layer in the TransGate encoder. In MetaCRS, we took 5 episodes in the exploration stage and 10 episodes in the conversational recommendation stage by default. 
We set $K_I$, $K_A$ and $K_{rec}$ to 10. We performed standard gradient decent in local adaptation with a learning rate of $0.01$, and updated the meta parameters using the Adam optimizer with a learning rate of $0.005$ and $L_2$ regularization coefficient 1e-6. The discount factor $\gamma$ is set to 0.999.  To make a fair comparison, we run 15 episodes in new user for adaptation in the second group of baselines. The size of query set is fixed to 10. The maximum turn $T$ in each episode is set to 10. We sample 5 users in each epoch when training MetaCRS. For all baselines, we used implementations provided by the papers and modified them as described before to support cold-start evaluation. 


\subsection{Overall Performance}
We report the comparison results across all methods in Table \ref{tab:overall}. We can clearly observe that MetaCRS outperformed all baselines with large margins. First of all, the results of the first group of baselines confirmed a single global policy cannot handle new users. By learning personalized policies, MetaCRS showed advantages in the final recommendation performance. 
In the second group of baselines, bandit-based algorithms select actions according to simple linear models, which are not capable to capture complicated relations between the algorithms' actions and user feedback, especially when positive rewards are discrete and sparse. Hence, such solutions performed much worse than other deep learning based methods. Both FPAN and UR can provide adaptive recommendations like \model{}, and thus they outperformed all baselines in the first group. Interestingly, we observe that FPAN outperformed EAR considerably. Different from EAR, FPAN updates user embeddings dynamically with users' positive and negative feedback on attributes and items by two gate modules, which enable dynamic item recommendation as in \model{}. This improved performance proves the necessity of adaptive recommendation during the course of CRS. UR showed general improvement against UNICORN and FPAN, both of which rely on a fixed FM model to recommend items (UNICORN also uses the FM model to pre-select items as actions) when performing policy training. Our state-based item recommender is able to provide improved recommendations at each time step by utilizing training signal from accepted items, which brings concrete benefits. 





In the third group, with the adaptation on new users, FT and IA led to general improvement, which indicates the necessity of policy adaptation in new users. Specifically, IA outperformed FT in most cases, which again proves that personalized polices are beneficial. Even though FPAN is able to dynamically capture users' preference with its gating modules, such knowledge did not generalize well on new users, which limited the improvement in F-FT and F-IA. By adapting both conversation and recommendation components in CRS, \model{} and UR-IA showed improvements against other baselines, which validates the necessity of our decoupled adaptation strategy. But as UR is not trained for generalization, it is hard to find optimal personalized policies for new users starting from such a global policy in UR-FT and UR-IA, while the meta model in \model{} is trained for generalization. In addition, the meta-exploration policy provides useful information for fast adaptation. Thus \model{} is able to perform better even with fewer adaptation episodes (first 5 episodes are used to explore user preferences).

\subsection{Ablation Study}
\subsubsection{Impact of support set size.} 
\begin{figure}[!htp]
\begin{subfigure}{0.235\textwidth}
  \centering
  \includegraphics[width=4.25cm]{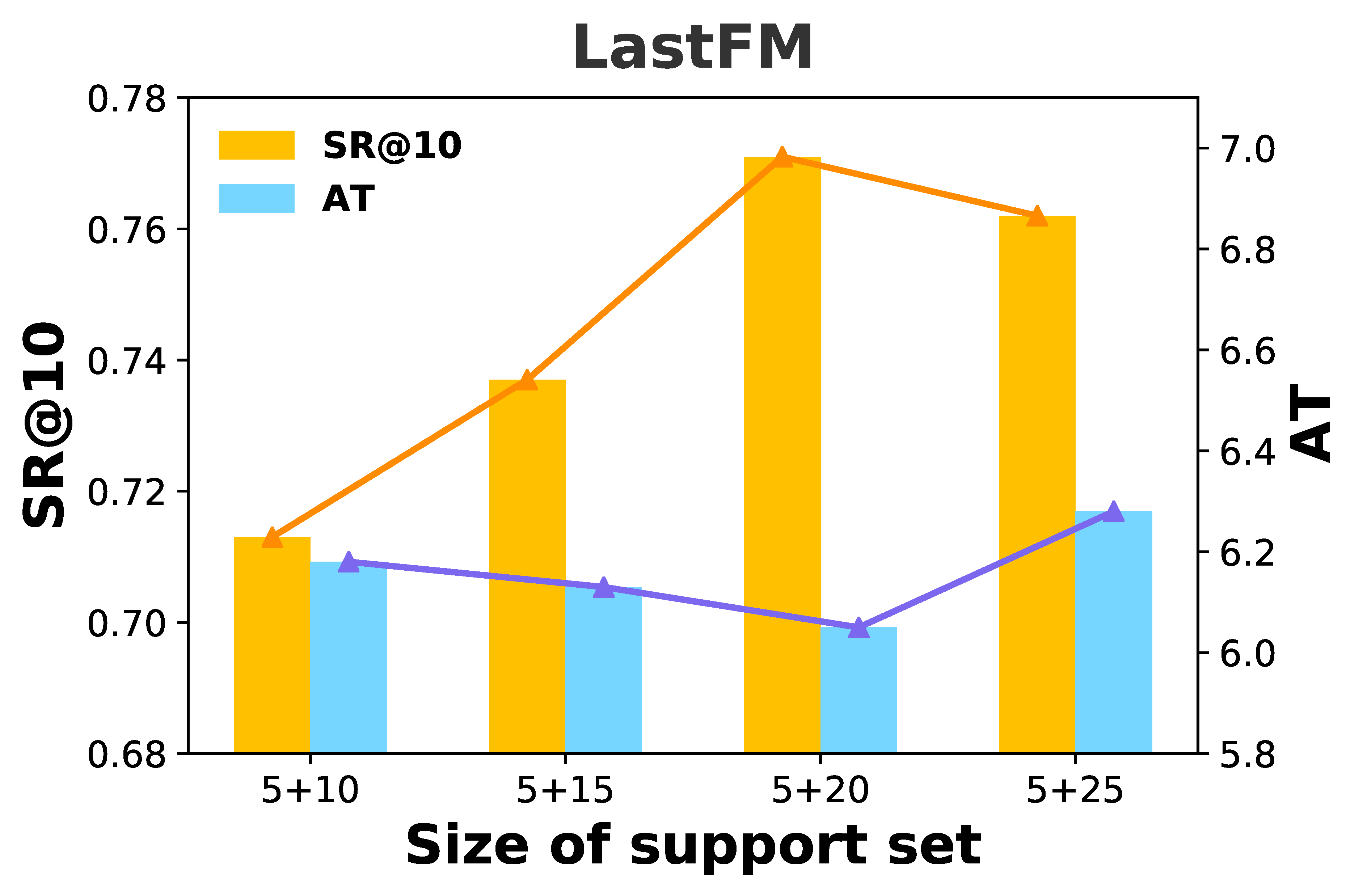}
\end{subfigure}
\begin{subfigure}{0.235\textwidth}
  \centering
  \includegraphics[width=4.25cm]{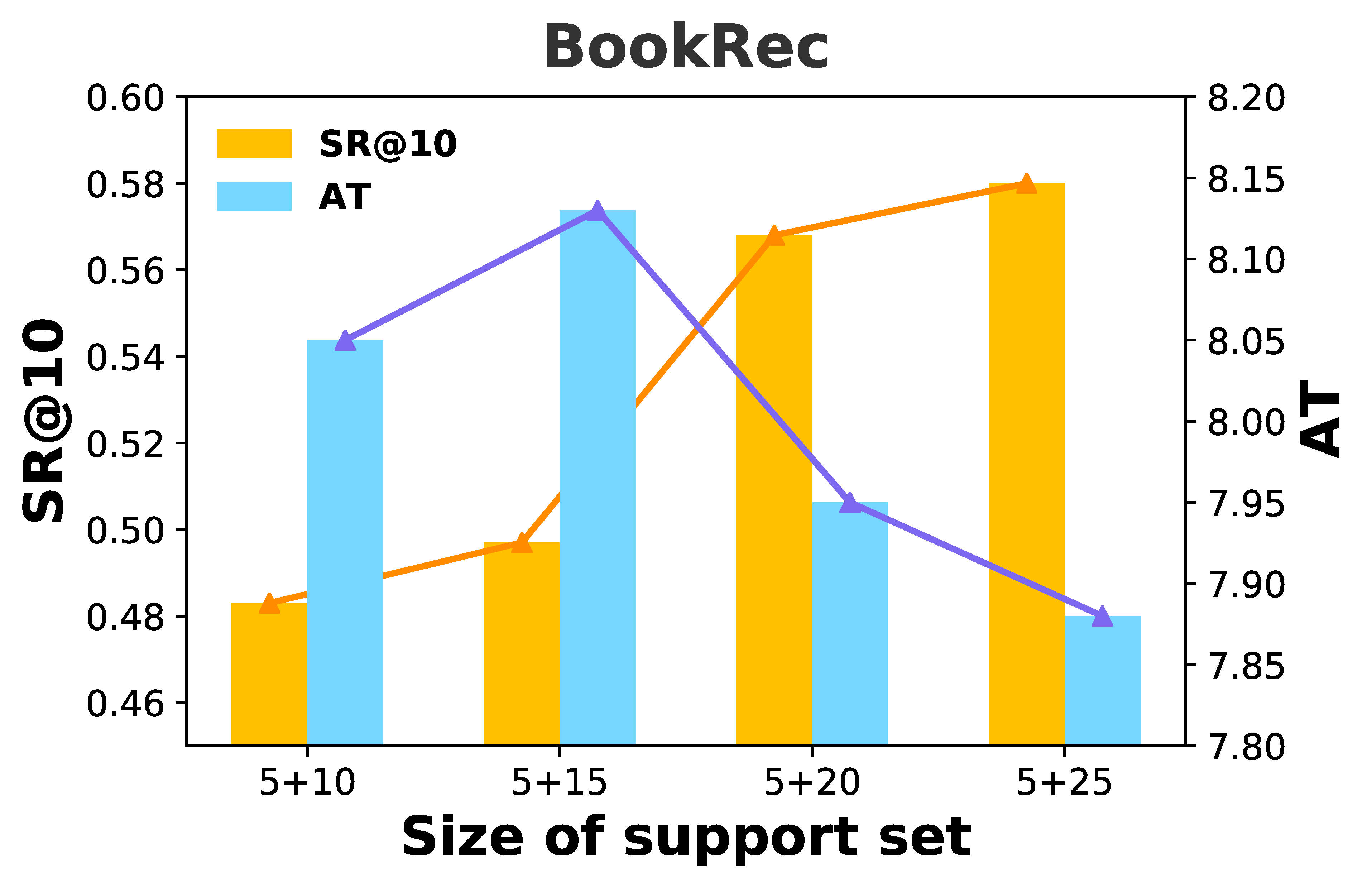}
\end{subfigure}
\caption{Performance comparisons w.r.t. size of support set.}
\vspace{-3mm}
\label{fig:support}
\end{figure}

Since policy adaptation is performed on the support set, it is important to study how many episodes are needed to obtain a good personalized policy (RQ3). In this experiment, we gradually increased the size of support sets with a step size 5. We kept the size of exploration episodes unchanged since 5 episodes are empirically sufficient for pinning down the target user's preference. Due to space limit, we only reported the results on the LastFM and BookRec dataset in Figure \ref{fig:support}, and similar results were also observed on the MovieLens dataset. With a larger support set, the success rate increases considerably and the number of average turn also reduces. This is expected since more observations can be collected to better adapt the meta policy for each user.
And this result also demonstrates the promise of personalized CRS policy learning: the quality of recommendation increases rapidly as the users get engaged with the system, which leads to a win-win situation for both users and system.

\subsubsection{Impact of different \model{} components.} In this section, we study the contribution of different components in \model{} to answer RQ4. Firstly, we evaluated the model's performance without local adaptation, which  essentially evaluated the learnt meta policy. 
Secondly, we removed the meta-exploration policy and directly executed policy adaptation. This setting shows how a dedicated exploration strategy affects policy adaptation. Finally, we replace the TransGate module with a linear layer similar to \cite{zhao2018recommendations} to study how state representation learning affects the CRS performance. In particular, the positive and negative embeddings are obtained by taking the average of all positive and negative feedback separately. 

\begin{table}[!htp]
\caption{Ablation analysis in \model{}.}
\vspace{-3mm}
\label{tab:ablation}
\begin{tabular}{c|cc|cc|cc}
\hline
 & \multicolumn{2}{c|}{LastFM} & \multicolumn{2}{c|}{BookRec} &           \multicolumn{2}{c}{MoiveLens} \\ \hline
              & SR@10 & AT    & SR@10 &  AT &  SR@10 & AT         \\ \hline
$\neg$adaptation  & 0.632 & 6.68 & 0.378 & 8.56 &  0.630 & 7.31     \\
$\neg$exploration &  0.677 &  6.64   &   0.411    &    8.37   &  0.738 &   6.65  \\
$\neg$TransGate  & 0.678 & 6.41  &  0.428  &   8.51    &  0.724 & 6.95            \\
MetaCRS   &  \textbf{0.713}  &  \textbf{6.18} &  \textbf{0.487}  & \textbf{8.06}    &  \textbf{0.745}  & \textbf{6.27}     \\  
\hline
\end{tabular}
\vspace{-1mm}
\end{table}

We present the results in Table \ref{tab:ablation}. Firstly, we can observe the performance before adaptation is not bad, or even better than most of our baselines in Table \ref{tab:overall}, which suggests the meta policy in \model{} already captured some important patterns for interacting with users. We can further compare the learnt meta policy with UR in Table \ref{tab:overall}, which shares the same state encoder and item recommender, but was trained globally. UR is slightly better than the meta policy in \model{}. The reason is UR is trained to maximize performance on training users and generalized by the i.i.d. assumption. But the meta policy is trained to maximize the adapted policies' performance, not its own performance on new users. Hence, when testing users share reasonable similarity with training users, UR can be effective in serving the testing users.
But we can observe a large performance gain after adaptation, which proves the meta policy successfully serves as a good starting point for fast adaptation. 
Next, it is clear that without the exploration stage the performance degenerates. It confirms recognizing who the system is serving is critical for a successful adaptation. 
We finally evaluate the effectiveness of TransGate, without which the performance degenerates on all three datasets. 
This demonstrates the necessity of fine-grained modeling of user feedback, especially the negative feedback, for understanding users' preferences.